\documentclass[11pt,letterpaper]{article}

\usepackage{times}
\usepackage{latexsym}
\usepackage{graphicx}
\usepackage{ulem}
\usepackage{url}
\usepackage{caption} 
\usepackage{amsmath}
\DeclareMathOperator*{\argmax}{arg\,max}
\usepackage{emnlp2017}
\emnlpfinalcopy

\def\emnlppaperid{***}

\title{Here's My Point: Joint Pointer Architecture for Argument Mining}

\author{Peter Potash, Alexey Romanov, Anna Rumshisky \\
		Department of Computer Science\\
        University of Massachusetts Lowell\\
  {\tt \{ppotash,aromanov,arum\}@cs.uml.edu}}

\date{}

\begin{document}

\maketitle

\begin{abstract}



In order to determine argument structure in text,
one must understand how different individual components of the
overall argument are linked.
This work presents the first neural network-based approach to link extraction in argument mining. 
Specifically, we propose a novel architecture that applies Pointer Network sequence-to-sequence attention modeling to structural prediction in discourse parsing tasks.
We then develop a joint model that extends this architecture to simultaneously address the link extraction task and the classification of argument components.  
The proposed joint model achieves state-of-the-art results on two separate evaluation corpora, showing far superior performance than the previously proposed corpus-specific and heavily feature-engineered models.
Furthermore, our results demonstrate that jointly optimizing for both tasks is crucial for high performance.
\end{abstract}

\section{Introduction}

An important goal in argument mining is to
understand the structure in argumentative text \cite{persing2016end,peldszus2015joint,stab2016parsing,nguyencontext}. One fundamental
assumption when working with argumentative text is the presence of
Arguments Components (ACs). The
types of ACs are generally characterized as a \textit{claim} or a
\textit{premise} \cite{govier2013practical}, with premises acting as
support (or possibly attack) units for claims (though some corpora have
further AC types, such as \textit{major claim}
\cite{stab2016parsing,stab2014identifying}).

\begin{figure*}
\begin{minipage}{0.4\textwidth}
First, [\uwave{cloning will be beneficial for many people who are in need of organ transplants}]$_{AC1}$. In addition, [\uline{it shortens the healing process}]$_{AC2}$. Usually, [\uline{it is very rare to find an appropriate organ donor}]$_{AC3}$
and [\uline{by using cloning in order to raise required organs the waiting time can be shortened tremendously}]$_{AC4}$.
\end{minipage} \hfill
\begin{minipage}{0.05\textwidth}
    \centering 
    \includegraphics[clip, trim=0.5cm 0.5cm 24cm 17.8cm, scale=0.5]{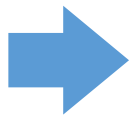}
\end{minipage} \hfill
\begin{minipage}{0.45\textwidth}


    \centering 
    \includegraphics[clip, trim=0.5cm 0.5cm 0.5cm 0.5cm, scale=0.6]{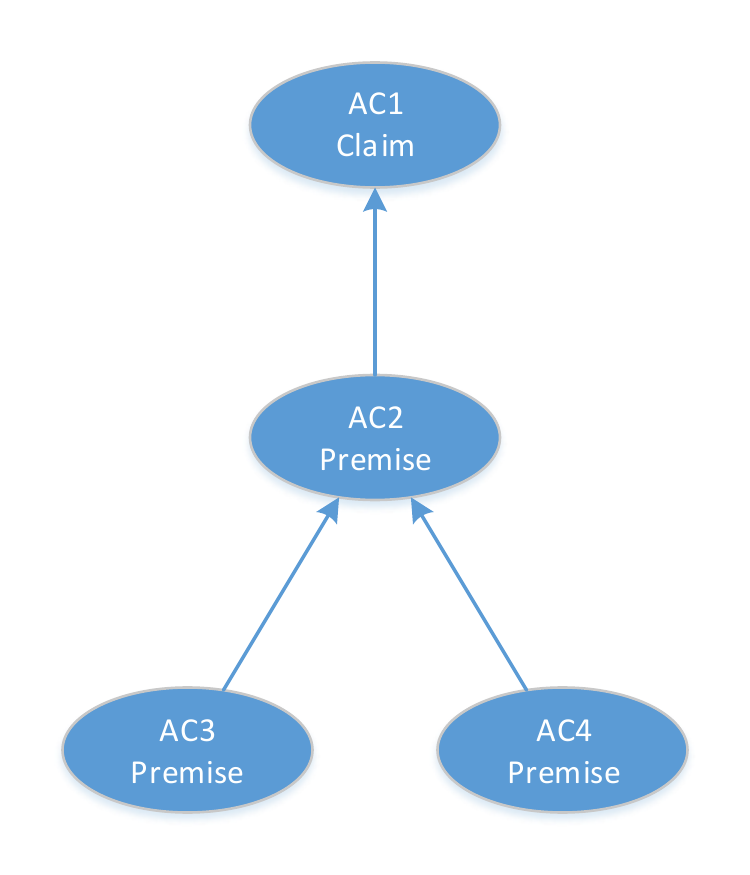}

\end{minipage}
    \caption{An example of argument structure with four ACs. The left
    side shows raw text that has been annotated for the presence of ACs.
    Squiggly and straight underlining means an AC is a \textit{claim} or
    \textit{premise}, respectively. The ACs in the text have also been
    annotated for links to other ACs, which is shown in the right
    figure. ACs 3 and 4 are
    \textit{premises} that link to another \textit{premise}, AC2. Finally,
    AC2 links to a \textit{claim}, AC1. AC1 therefore acts as the central
    argumentative component.}
    \label{fig:argument_structure} 
\end{figure*}

The task of processing argument structure encapsulates four
distinct subtasks
(our work focuses on
subtasks 2 and 3): (1) Given a sequence of tokens that represents an
entire argumentative text, determine the token subsequences that constitute
non-intersecting ACs; (2) Given an AC, determine the type of AC (\textit{claim},
\textit{premise}, etc.); (3)
Given a set/list of ACs, determine which ACs have directed links that encapsulate
overall argument structure; (4) Given two linked ACs, determine whether the
link is a supporting or attacking relation.
This can be labeled as a
`micro' approach to argument mining \cite{stab2016parsing}.
In contrast,
there have been a number of efforts to identify argument structure at a higher level
\cite{boltuzic2014back,ghosh2014analyzing,cabrio2012natural}, as well as slightly
re-ordering the pipeline with respect to AC types \cite{rinott2015show}).

There are two key assumptions our work makes going forward. First, we assume
subtask 1 has been completed, i.e. ACs have already been identified. Second,
we follow previous work that assumes a tree structure
for the linking of ACs 
\cite{palau2009argumentation,cohen1987analyzing,peldszus2015joint,stab2016parsing}. 
Specifically, a given AC can only have a single 
outgoing link, but can
have numerous incoming links. Furthermore, there is a `head' component that
has no outgoing link (the top of the tree). 
Depending on the corpus (see Section \ref{sec:exp_des}), an argument
structure can be either a single tree or a forest, consisting of multiple trees.
Figure \ref{fig:argument_structure} shows
an example that we will use throughout the paper to concretely explain how our approach
works. First, the left side of the figure presents the raw text of a paragraph in a 
persuasive essay \cite{stab2016parsing}, with the ACs contained in square brackets.
Squiggly
verse straight underlining differentiates between claims and premises, respectively.
The ACs have been annotated as to how they are linked, and the right side of the figure
reflects this structure.
The
argument structure with four ACs forms a tree, where AC2 has
two incoming links, and AC1 acts as the head, with no outgoing links.
We also specify the \textit{type} of AC, with the head AC marked as a
\textit{claim} and the remaining ACs marked as \textit{premises}.
Lastly, we note that the order of argument components can be a strong
indicator of how components should related. Linking to the first argument
component can provide a competitive baseline heuristic \cite{peldszus2015joint,stab2016parsing}.

Given the above considerations, we propose that sequence-to-sequence 
attention modeling, in the spirit of a Pointer Network (PN) \cite{vinyals2015pointer}, 
can be effective for predicting argument structure.
To the best of our knowledge, a clean, elegant implementation of a PN-based model has yet to be introduced for discourse parsing tasks.
A PN is a sequence-to-sequence model \cite{sutskever2014sequence} that outputs a distribution over the encoding indices at each decoding timestep. 
More generally, it is a recurrent model with attention \cite{bahdanau2014neural}, and we claim that as such, it is promising for link extraction because it inherently possesses three important characteristics: (1) it is able to model the sequential nature of
ACs, (2) it constrains ACs to have a single outgoing link, thus partly enforcing the
tree structure, and (3) the hidden representations learned by the model can be
used for jointly predicting multiple subtasks. 
Furthermore, we believe the sequence-to-sequence aspect of the model provides two
distinct benefits: (1) it allows for two separate representations of a single AC (one for the source and one for the target of the link), and 
(2) the decoder LSTM
could learn to predict correct 
sequences of linked indices, which is a second recurrence over ACs.
Note that we also test the sequence-to-sequence architecture against a simplified model that only uses hidden states from an encoding network to make predictions (see Section \ref{sec:res}).

The main technical contribution of our work is a joint model that
simultaneously predicts links between ACs and determines their \textit{type}.
Our joint model uses the hidden representation of ACs produced during 
the encoding step (see Section \ref{sec:jnm}). 
While PNs were originally proposed to allow a variable length decoding
sequence, our model differs in that it decodes for the same number of 
timesteps as there are inputs.
This is a key insight that allows for a sequence-to-sequence to be used
for structural prediction.
Aside from the partial assumption of tree structure in the argumentative
text, our models do not make any additional assumptions about the AC types or 
connectivity, unlike the work of \newcite{peldszus2014towards}. 
Lastly, in respect to the broad task of parsing, our model is
flexible because it can easily handle non-projective, multi-root dependencies.
We evaluate our models on the corpora of \newcite{stab2016parsing} and 
\newcite{peldszus2014towards}, and compare our results with the results
of the aformentioned authors. Our results show that (1) joint modeling is imperative for
competitive performance on the link extraction task, (2) the presence of the second
recurrence improves performance over a non-sequence-to-sequence model, and 
(3) the joint model can outperform models with heavy feature-engineering
and corpus-specific constraints.

\section{Related Work}

Palau and Moens \shortcite{palau2009argumentation} is an early work in 
argument mining,
using a hand-crafted Context-Free Grammar to determine the structure
of ACs in a corpus of legal texts. Lawrence et al. \shortcite{lawrence2014mining}
leverage a topic modeling-based AC similarity to uncover tree-structured
arguments in philosophical texts. Recent work
offers data-driven approaches to the task of predicting links between ACs.
Stab and Gurevych \shortcite{stab2014identifying} approach the task as a
binary classification problem. The authors train an SVM with various
semantic and structural features. Peldszus and Stede have also 
used classification models for predicting the presence of links 
\shortcite{peldszus2015joint}.
The first neural network-based model for argumentation mining was proposed by
\cite{lahaempirical}, who use two recurrent networks in end-to-end fashion to
classify AC types.

Various authors have also proposed to
jointly model link extraction with other subtasks from the argumentation
mining pipeline, using either a Integer Linear Programming (ILP) framework
\cite{persing2016end,stab2016parsing} or directly feeding previous subtask
predictions into a tree-based parser. The former joint approaches are evaluated
on an annotated corpus of persuasive essays \cite{stab2014annotating,stab2016parsing},
and the latter on a corpus of microtexts \cite{peldszus2014towards}.
The ILP framework is effective in enforcing a tree structure between ACs
when predictions are made from otherwise naive base classifiers.

Recurrent neural networks have
previously been proposed to model tree/graph structures in a linear manner.
\cite{vinyals2015grammar} use a sequence-to-sequence
model for the task of syntactic parsing.
\cite{bowman2015tree} experiment on an artificial entailment
dataset that is specifically engineered to capture recursive logic
\cite{bowman2014recursive}.
Standard recurrent
neural networks can take in complete sentence sequences
and perform competitively with a recursive neural network.
Multi-task learning for sequence-to-sequence has also been proposed
\cite{luong2015multi},
though none of the 
models used a PN for prediction.

In the field of discourse parsing, the work of Li et al. 
\shortcite{lidiscourse} is the only work, to our knowledge, 
that incorporates attention into the network architecture.
However, the attention is only used in the process of creating representations of the text itself. Attention is
\textit{not} used to predict the overall discourse structure.
In fact, the model still relies on a binary classifier to
determine if textual components should have a link.
Arguably the most similar approach to ours is in the field of dependency 
parsing \cite{cheng2016bi}. The authors
propose a model that performs `queries' between word representations
in order to determine a distribution over potential headwords.

\begin{figure*}[ht]
    \centering 
    \includegraphics[clip, trim=3.8cm 8cm 3cm 8cm, scale=0.6]{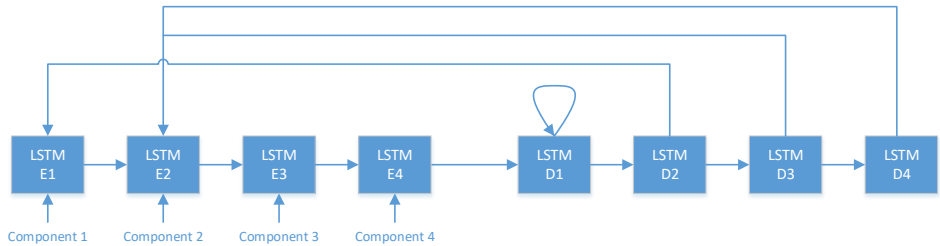}
    \caption{Applying a Pointer Network to the example paragraph in Figure~\ref{fig:argument_structure} with LSTMs unrolled over time. Note that D1 points to itself to denote that it
    has not outgoing link and is therefore the head of
    a tree.}
    \label{fig:model_regular} 
\end{figure*}

\section{Proposed Approach}\label{sec:pn_model}

In this section, we describe our approach to using a sequence to-sequence model with attention for argument mining, specifically, identifying AC types and extracting the links between them.  We begin by giving a brief overview of these models.


\subsection{Pointer Network}

A PN is a sequence-to-sequence model \cite{sutskever2014sequence} with
attention \cite{bahdanau2014neural} that was proposed to handle decoding
sequences over the encoding inputs, and can be extended to arbitrary
sets \cite{vinyals2015order}. The original
motivation for a pointer network was to allow networks to learn solutions
to algorithmic problems, such as the traveling salesperson and convex hull problems,
where the solution is a sequence over input points. The PN model is
trained on input/output sequence pairs $(E,D)$, where $E$ is the source and
$D$ is the target (our choice of $E$,$D$ is meant to represent the encoding,
decoding steps of the sequence-to-sequence model). Given model parameters
$\Theta$, we apply the chain rule to determine the probability of a single
training example:
\begin{equation}\label{eq:tr_prob}
p(D|E;\Theta) = \prod_{i=1}^{m(E)}p(D_i|D_1,...,D_{i-1},E;\Theta)
\end{equation}
\noindent where the function $m$ signifies that the number of decoding timesteps is a
function of each individual training example. We will discuss shortly why we
need to modify the original definition of $m$ for our application.
By taking the log-likelihood of Equation \ref{eq:tr_prob}, we arrive at the
optimization
objective:
\begin{equation}\label{eq:opt_obj}
\Theta^{*} = \argmax_{\Theta}\sum_{E,D}\log p(D|E;\Theta)
\end{equation}
which is the sum over all training example pairs.

The PN uses Long Short-Term Memory (LSTM) \cite{hochreiter1997long} for
sequential modeling, which produces a hidden layer $h$ at each
encoding/decoding timestep. In practice, the PN has two separate LSTMs, one
for encoding and one for decoding. Thus, we refer to encoding hidden layers
as $e$, and decoding hidden layers as $d$.

The PN uses a form of content-based attention \cite{bahdanau2014neural} to
allow the model to produce a distribution over input elements. This can also
be thought of as a distribution over input indices, wherein a decoding step
`points' to the input. Formally, given encoding hidden states $(e_1,...,e_n)$,
the model calculates $p(D_i|D_1,...,D_{i-1},E)$ as follows:
\begin{equation}\label{eq:calc_u}
u_j^i = v^T \tanh (W_1 e_j + W_2 d_i)
\end{equation}
\begin{equation}\label{eq:calc_dist}
p(D_i|D_1,...,D_{j-1},E) = softmax(u^i)
\end{equation}
\noindent where matrices $W_1$, $W_2$ and vector $v$ are parameters of the model
(along with the LSTM parameters used for encoding and decoding).
In Equation \ref{eq:calc_u}, prior to taking the dot product with $v$, the
resulting transformation can be thought of as creating a joint hidden
representation of inputs $i$ and $j$.
Vector
$u^i$ in equation \ref{eq:calc_dist} is of length $n$, and index $j$ corresponds to input element $j$. Therefore, by taking the softmax of $u^i$,
we are able to create a distribution over the input.

\subsection{Link Extraction as Sequence Modeling}

\begin{figure*}[ht]
    \centering 
    \includegraphics[clip, trim=3.2cm 5.8cm 2cm 5.6cm, scale=0.6]{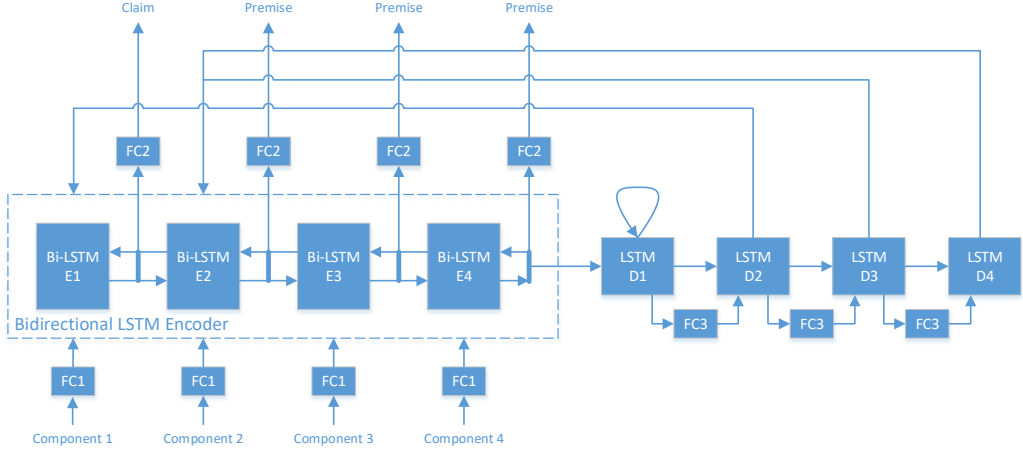}
    \caption{Architecture of the joint model applied to the example in
    Figure~\ref{fig:argument_structure}.
    Note that D1 points to itself to denote that it
    has not outgoing link and is therefore the head of
    a tree.}
    \label{fig:model_joint} 
\end{figure*}

A given piece of text has a set of ACs, which occur in a
specific order in the text: $(C_1,...,C_n)$. Therefore, at encoding 
timestep $i$, the model is fed a representation of $C_i$. 
Since the
representation is large and sparse (see Section \ref{sec:rep_ac} for details on how we
represent ACs), we add a fully-connected layer before
the LSTM input. 
Given a representation $R_i$ for AC $C_i$, the LSTM input
$A_i$ is calculated as:
\begin{equation}\label{eq:fc_input}
A_i = \sigma(W_{rep}R_i + b_{rep})
\end{equation}
\noindent where $W_{rep}$, $b_{rep}$ in turn become model parameters, and $\sigma$
is the sigmoid function\footnote{We also experimented with relu and elu activations,
but found sigmoid to yield the best performance.}. Similarly, the decoding network
applies a
fully-connected layer with sigmoid activation to its inputs, see Figure 
\ref{fig:model_joint}. At 
encoding step $i$, the encoding LSTM produces hidden layer $e_i$, which can
be thought of as a hidden representation of AC $C_i$.

In order to make sequence-to-sequence modeling applicable to the problem of link extraction, we
explicitly set the number of decoding timesteps to be equal to the number of
input components. Using notation from Equation \ref{eq:tr_prob}, the
decoding sequence length for an encoding sequence $E$ is 
simply $m(E) = |\{C_1,...,C_n\}|$, which is trivially equal to $n$. By constructing the decoding
sequence in this manner, we can associate decoding timestep $i$ with AC
$C_i$.

From Equation \ref{eq:calc_dist}, decoding timestep $i$ will output a
distribution over input indices. The result of this distribution will
indicate to which AC component $C_i$ links. Recall there is a possibility that
an AC has no outgoing link, such as if it's the root of the
tree. In this case, we state that if AC $C_i$ does not have an outgoing link,
decoding step $D_i$ will output index $i$. Conversely, if $D_i$ outputs
index $j$, such that $j$ is not equal to $i$, this implies that $C_i$ has an
outgoing link to $C_j$. For the argument structure in Figure 
\ref{fig:argument_structure}, the corresponding decoding sequence is $(1,1,2,2)$.
The topology of this decoding sequence is illustrated in Figure \ref{fig:model_regular}. Observe
how $C_1$ points to itself since it has no outgoing link.

Finally, we note that we have a Bidirectional LSTM \cite{graves2005framewise} as the
encoder, unlike the model proposed by
\newcite{vinyals2015pointer}. Thus, $e_i$ is the concatenation of forward and backward hidden
states $\overrightarrow{e}_i$ and $\overleftarrow{e}_{n-i+1}$, produced by two
separate LSTMs. The decoder remains a standard forward LSTM.

\subsection{Representing Argument Components}
\label{sec:rep_ac}

At each timestep of the encoder, the network takes in a representation of an AC.
Each AC is itself a sequence of tokens, similar to the
Question-Answering dataset from \newcite{weston2015towards}. We follow the work of
\newcite{stab2016parsing} and focus on three different types of features
to represent our ACs: (1) Bag-of-Words of the AC; (2) Embedding representation based on GloVe embeddings \cite{pennington2014glove}, which uses average, max, and min pooling across the token embeddings; (3) Structural features: Whether or not the AC is the first AC in a paragraph, and Whether the AC is in an opening,
body, or closing paragraph. 
See Section \ref{sec:dis} for an ablation study of the proposed features.

\subsection{Joint Neural Model}
\label{sec:jnm}

Up to this point, we focused on the task of extracting links between ACs. However, recent work has shown that joint models that simultaneously try to complete multiple aspects of the subtask
pipeline outperform models that focus on a single subtask
\cite{persing2016end,stab2014identifying,peldszus2015joint}.
Therefore, we will modify the single-task architecture so that it
would allow us to perform AC classification \cite{kwon2007identifying,rooney2012applying} together with
link prediction. Knowledge of an individual subtask's predictions can aid in other subtasks. For example, \textit{claims} do not have an outgoing link, so knowing the type of AC can aid in the link prediction task. This can be seen as a way of regularizing the hidden representations from the encoding component \cite{che2015deep}.

At each timestep, predicting AC type is a straightforward classification task: given AC $C_i$, we need to predict whether it is a \textit{claim}, \textit{premise}, or possibly \textit{major claim}.
More generally, this is another sequence modeling problem: given input
sequence $E$, we want to predict a sequence of argument types $T$.
For encoding timestep $i$, the model creates hidden representation $e_i$.
This can be thought of as a representation of AC $C_i$. Therefore, our joint model will simply pass this representation through a fully-connected layer as follows:
\begin{equation}
z_i = W_{cls}e_i + b_{cls}
\end{equation}
\noindent where $W_{cls}$, $b_{cls}$ become elements of the model
parameters, $\Theta$.
The dimensionality of $W_{cls}$, $b_{cls}$ is determined by the number of classes. Lastly, we use softmax to form a distribution over the possible classes.

Consequently, the
probability of predicting the component type at timestep $i$ is defined as:
\begin{equation}
p(T_i|E_i;\Theta) = softmax(z_i)
\end{equation}
\noindent Finally, combining this new prediction task with Equation 
\ref{eq:opt_obj},
we arrive at the new training objective:
\begin{equation}
\begin{split}
\Theta^* = \argmax_{\Theta}\alpha\sum_{E,D}\log p(D|E;\Theta)\\+
(1-\alpha)\sum_E \log
p(T|E;\Theta)
\end{split}
\end{equation}
\noindent which simply sums the costs of the individual prediction tasks,
and the second summation is the cost for the new task of predicting AC type. 
$\alpha \in [0,1]$ is a hyperparameter that specifies how we weight
the two prediction tasks in our cost function.
The architecture of the joint model, applied to our ongoing example, is illustrated in Figure \ref{fig:model_joint}.

\section{Experimental Design}
\label{sec:exp_des}


As we have mentioned, 
our work assumes that ACs have already been identified.
The order of ACs corresponds directly to the
order in which the ACs appear in the text. 
We test the effectiveness of our proposed model on a dataset of
persuasive essays (PEC)
\cite{stab2016parsing},
as well as a dataset of microtexts (MTC)
\cite{peldszus2014towards}.
The feature space for the PEC
has roughly 3,000 dimensions, and the MTC feature space has between
2,500 and 3,000 dimensions, depending on the data split. The PEC contains a total of
402 essays, with a frozen set of 80 essays held out for testing. There are
three AC types in this corpus: \textit{major claim}, \textit{claim},
and \textit{premise}.
In this corpus individual structures can be either trees or forests.
Also, in this corpus, each essay has multiple paragraphs,
and argument structure is only uncovered
within a given paragraphs.
The MTC contains 112 short texts. Unlike the PEC,
each text in this corpus is itself a complete example, as well as a single tree. Since the
dataset is small, the authors have created 10 sets of 5-fold cross-validation,
reporting the
the average across all splits for final model evaluation. This corpus 
contains only two types of ACs:
\textit{claim} and \textit{premise}.
Note that link prediction is directed, i.e., predicting a link between the pair $C_i,C_j (i \neq j)$ is different than $C_j,C_i$.

\begin{table*}[ht]
\centering
\begin{tabular}{|l||c|c|c|c||c|c|c|}
\hline
  & \multicolumn{4}{c||}{Type prediction} & \multicolumn{3}{c|}{Link prediction}  \\ \hline
 Model & Macro f1 & MC f1 & Cl f1 & Pr f1 & Macro f1 & Link f1 & No Link f1 \\ \hline
 Base Classifier & .794 & .891  & .611 & .879 & .717 & .508 &  .917 \\ \hline
 ILP Joint Model & .826 & .891 & .682 & .903 & .751 & .585 &  .918 \\ \hline \hline
 Single-Task Model & - & - & - & - & .709 & .511 & .906 \\ \hline
 Joint Model No Seq2Seq & .810 & .830 & .688 & .912 & .754 & .589 & .919 \\ \hline
 Joint Model No FC Input & .791 & .826 & .642 & .906 & .708 & .514 & .901 \\ \hline
 Joint Model & \bf .849 & \bf .894 & \bf .732 & \bf .921 & \bf .767 & \bf .608 & \bf .925 \\ \hline
\end{tabular}
\caption{Results on the Persuasive Essay corpus. All models we tested are joint models,
except for the Single-Task Model model, which only predicts links. All model have a fully-connected
input layer, except for the row titled `Joint Model No FC Input'. See Section
\ref{sec:res} for a full description of the models.}
\label{tbl:pe_res}
\end{table*}

\begin{table*}[ht]
\centering
\begin{tabular}{|l||c|c|c||c|c|c|}
\hline
  & \multicolumn{3}{c||}{Type prediction} & \multicolumn{3}{c|}{Link prediction}  \\ \hline
 Model & Macro f1 & Cl f1 & Pr f1 & Macro f1 & Link f1 & No Link f1 \\ \hline
 Simple & .817 & - & - & .663 & .478 & .848 \\ \hline
 Best EG & \bf .869 & - & - & .693 & .502 & .884 \\ \hline
 MP+p & .831 & - & - & .720 & .546 & .894 \\ \hline 
 Base Classifier & .830 &  .712 & .937 & .650 & .446 &  .841 \\ \hline
 ILP Joint Model & .857 &  .770 & .943 & .683 & .486 & .881 \\ \hline \hline
Joint Model & .813 & .692 & .934 & \bf .740 & \bf  .577 & \bf .903 \\ \hline
\end{tabular}
\caption{Results on the Microtext corpus.}
\label{tbl:mt_res}
\end{table*}

We implement our models in TensorFlow
\cite{tensorflow2015-whitepaper}. We use the following parameters:
hidden input dimension size 512, hidden layer size 256 for the bidirectional LSTMs,
hidden layer size 512 for the LSTM decoder, $\alpha$ equal to 0.5, and dropout
\cite{srivastava2014dropout}
of 0.9. We believe the need for such high dropout is due to the small amounts of
training data \cite{zarrella2016mitre}, particularly in the MTC.
All models are
trained with Adam optimizer \cite{kingma2014adam} with a batch size of 16. For
a given training set, we randomly select 10\% to become the validation set.
Training occurs for 4,000 epochs. Once training is completed, we select the model
with the
highest validation accuracy (on the link prediction task)
and evaluate it on the held-out test set. 
At test time, we take a greedy approach and select the index of the probability distribution (whether link or type prediction) with the highest value.

\section{Results}
\label{sec:res}

\begin{table*}[ht]
\centering
\begin{tabular}{|l||c|c|c|c||c|c|c|}
\hline
  & \multicolumn{4}{c||}{Type prediction} & \multicolumn{3}{c|}{Link prediction}  \\ \hline
 Model & Macro f1 & MC f1 & Cl f1 & Pr f1 & Macro f1 & Link f1 & No Link f1 \\ \hline
 No structural & .808 & .824  & .694 & .907 & .760 & .598 &  .922 \\ \hline
 No BOW & .796 & .833 & .652 & .902 & .728 & .543 &  .912 \\ \hline
 No Embeddings & .827 & .874 & .695 & .911 & .750 & .581 & .918 \\ \hline
 Only Avg Emb* & .832 & .873 & .717 & .917 & .751 & .583 & .918 \\ \hline
 Only Max Emb* & .843 & .874 & \bf .732 & \bf .923 & .766 & \bf .608 & .924 \\ \hline
 Only Min Emb* & .838 & .878 & .719 & .918 & .763 & .602 & .924 \\ \hline
 All features & \bf .849 & \bf .894 & \bf .732 & .921 & \bf .767 & \bf .608 & \bf .925 \\ \hline
\end{tabular}
\caption{Feature ablation study. * indicates that both BOW and Structural are present,
as well as the stated embedding type.}
\label{tbl:abl_study}
\end{table*}

\begin{table*}
\centering
\begin{tabular}{|c||c|c|c|c||c|c|c|}
\hline
 & \multicolumn{4}{c||}{Type prediction} & \multicolumn{3}{c|}{Link prediction}  \\ \hline
Bin & Macro f1 & MC f1 & Cl f1 & Pr f1 & Macro f1 & Link f1 & No Link f1 \\ \hline
$1\leq len < 4$ & .863 & .902 & .798 & .889 & .918 & .866 & .969 \\ \hline
$4\leq len < 8$ & .680 & .444 & .675 & .920 & .749 & .586 & .912\\ \hline
$8\leq len < 12$ & .862* & .000* & .762 & .961 & .742 & .542 & .941 \\ \hline
\end{tabular}
\caption{Results of binning test data by length of AC sequence. * indicates
that this bin does not contain any \textit{major claim} labels, and this average
only applies to \textit{claim} and \textit{premise} classes. However, we do not
disable the model from predicting this class: the model was able to avoid
predicting this class on its own.}
\label{tbl:len_bin}
\end{table*}

The results of our experiments are presented in Tables \ref{tbl:pe_res} and \ref{tbl:mt_res}. For each corpus, we present f1 scores for the AC type
classification experiment, with a macro-averaged score of the individual 
class f1 scores. We also present the f1 scores for predicting the 
presence/absence of links between ACs, as well as the associated macro-average
between these two values. 

We implement and compare four types of neural models: 1) The previously
described
joint model described in Section \ref{sec:jnm} (called Joint Model in the
tables); 2) The same as 1), but without the fully-connected input layers (called Joint Model No FC Input in the table);
3) The same as 1), but the model only predicts the link task, and is therefore
not optimized for type prediction (called Single-Task Model in the
tables);
4) A non-sequence-to-sequence model that uses the hidden layers produced
by the BLSTM encoder with the same type of attention as the joint
model (called Joint Model No Seq2Seq
in the table). That is, $d_i$ in Equation \ref{eq:calc_u} is replaced by $e_i$.

In both corpora we compare against the following previously proposed models:
Base Classifier \cite{stab2016parsing} is a feature-rich, task-specific (AC type 
or link extraction) SVM classifier. Neither of these classifiers enforce 
structural or global constraints. Conversely, the ILP Joint Model 
\cite{stab2016parsing} provides constraints by sharing prediction information
between the base classifiers. For example, the model attempts to enforce a tree
structure among ACs within a given paragraph, as well as using incoming link
predictions to better predict the type class \textit{claim}. For the MTC only, we also have the following comparative models: Simple 
\cite{peldszus2015joint} is a feature-rich logistic regression classifier. Best EG
\cite{peldszus2015joint} creates an Evidence Graph (EG) from the predictions of a 
set of base classifiers. The EG models the potential argument structure, and offers
a global optimization objective that the base classifiers attempt to optimize by adjusting their individual weights. 
Lastly, MP+p \cite{peldszus2015joint} combines predictions from base classifiers with a Minimum Spanning Tree Parser (MSTParser).

\section{Discussion}
\label{sec:dis}


First, we point out that the joint model achieves state-of-the-art on 10 of the
13 metrics in Tables \ref{tbl:pe_res} and \ref{tbl:mt_res}, including the
highest results in all metrics on the PEC, as well as
link prediction on the MTC. The performance on the MTC is very encouraging for several reasons. First, the fact that the model
can
perform so well with only a hundred training examples is rather remarkable.
Second, although we motivate the use of an attention model due to the fact that it partially
enforces a tree structure, other models we compare against explicitly contain
further constraints (for example, only premises can have outgoing links). Moreover, the MP+p model directly enforces the single tree
constraint unique to the microtext corpus (the PEC allows forests). Even though the joint model does not have the
tree constraint directly encoded, it able to learn the structure effectively from the training
examples so that it can outperform the Mp+p model for link prediction.
As for the other neural models, the joint model no seq2seq
performs competitively with the ILP joint model on the PEC, but trails the performance of the joint model. We believe this is
because the joint model is able to create two different representations for
each AC, one each in the encoding/decoding state, which benefits performance
in the two tasks. We also believe that the joint model benefits from a second
recurrence over the ACs, modeling the tree/forest structure in a linear manner.
Conversely, the joint model no seq2seq must encode information relating
to type as well as link prediction in a single hidden representation. On one hand,
the joint model no seq2seq outperforms the ILP model on link prediction, yet it is not able to
match the ILP joint model's performance on type prediction, primarily due to the
its poor performance on predicting the \textit{major claim} class. Another interesting outcome is the importance of the fully-connected layer before the LSTM input. This extra layer seems to be crucial for improving performance on this task. 
The results dictate that even a simple fully-connected layer with sigmoid activation can provide a useful dimensionality reduction step.
Finally, and arguably most importantly, the single-task model, only optimized for link prediction, suffers a large drop in performance, conveying that the dual optimization of the joint model is vital for high performance in the link prediction task.
We believe this is because the joint optimization creates more expressive representations of the ACs, which captures the natural relation between AC type and AC linking.

Table \ref{tbl:abl_study} shows the results of an ablation study for AC feature
representation. Regarding link prediction, BOW features are clearly the most
important, as their absence results in the highest drop in performance.
Conversely,
the presence of structural features provides the smallest
boost in performance, as the model is still able to record state-of-the-art results
compared to the ILP Joint Model.
This shows that the Joint Model is able to capture structural cues
through sequence modeling and semantics.
When considering type prediction, both BOW and
structural features are important, and it is the embedding features that provide
the least benefit. The Ablation results also provide an interesting insight into
the effectiveness of different pooling strategies for using individual token
embeddings to create a multi-word embedding. The popular method of
averaging embeddings (which is used by \newcite{stab2016parsing} in their
system) is in fact the worst method, although
its performance is still competitive with the previous state-of-the-art.
Conversely, max pooling results are on par with the joint model results in
Table \ref{tbl:pe_res}.

Table \ref{tbl:len_bin} shows the results on the PEC test set with
the test examples binned by sequence length. First, it is not surprising to see that the model
performs best when the sequences are the shortest (for link prediction; type prediction actually sees the worst performance in the middle bin). As the sequence length increases,
the accuracy on link prediction drops. This is possibly due to the fact that as the
length increases, a given AC has more possibilities as to which other AC it can link to,
making the task more difficult. Conversely, there is actually a rise in no link
prediction accuracy from the second to third row. This is likely due to the fact that
since the model predicts at most one outgoing link, it indirectly predicts no link
for the remaining ACs in the sequence. Since the chance probability is low for having
a link between a given AC in a long sequence, the no link performance is actually
better in longer sequences.

\section{Conclusion}

In this paper we have proposed how to use a joint sequence-to-sequence
model with attention \cite{vinyals2015pointer} to both 
extract links between
ACs and classify AC type. We evaluate our models on two corpora:
a corpus of persuasive essays \cite{stab2016parsing}, and a corpus of microtexts
\cite{peldszus2014towards}. The Joint Model
records state-of-the-art results on the persuasive essay corpus,
as well as achieving state-of-the-art results for link prediction on the microtext corpus.
The results show that jointly modeling the two prediction tasks is critical
for high performance.
Future work can attempt to learn the AC representations themselves, such as in
\newcite{kumar2015ask}.
Lastly, future work can integrate subtasks 1 and 4 into the model.
The representations produced by Equation \ref{eq:calc_u} could potentially be used to predict
link type, i.e. supporting or attacking (the fourth
subtask in the pipeline). In addition, a segmenting technique, such as the one proposed by
\newcite{weston2014memory}, can accomplish subtask 1.




\bibliography{emnlp2017}
\bibliographystyle{emnlp_natbib}

\end{document}